\documentclass[11pt,a4paper]{article}
\usepackage{emnlp2017}
\usepackage{times}
\usepackage{latexsym}

\usepackage{booktabs}
\usepackage{xspace}
\usepackage{stmaryrd}
\usepackage{amsmath}
\usepackage{amssymb}
\usepackage{amsthm}
\usepackage{multirow}
\usepackage{tikz}
\usepackage[small]{caption}
\usepackage{enumitem}

\emnlpfinalcopy

\def\emnlppaperid{300}

\DeclareMathOperator*{\argmin}{arg\,min}

\newcommand*\rot{\rotatebox{90}}

\newcommand{\world}{W}
\newcommand{\target}{X}
\newcommand{\nl}{e}
\newcommand{\rl}{f}
\newcommand{\denote}[1]{{\llbracket #1 \rrbracket}}
\newcommand{\rep}[1]{\textit{rep}(#1)}

\newcommand{\genx}{\textsc{GenX}\xspace}

\title{Analogs of Linguistic Structure in Deep Representations}

\author{Jacob Andreas \and Dan Klein \\
  Computer Science Division \\
  University of California, Berkeley \\
  {\tt {jda,klein}@cs.berkeley.edu}}

\date{}

\begin{document}

\maketitle

\begin{abstract}
We investigate the compositional structure of message vectors computed by a deep
network trained on a communication game. By comparing truth-conditional
representations of encoder-produced message vectors to human-produced referring
expressions, we are able to identify aligned (vector, utterance) pairs with the
same meaning. We then search for structured relationships among these aligned
pairs to discover simple vector space transformations corresponding to negation,
conjunction, and disjunction. Our results suggest that neural representations
are capable of spontaneously developing a ``syntax'' with functional analogues
to qualitative properties of natural language.\footnote{ Code and data are
available at \url{http://github.com/jacobandreas/rnn-syn}.}
\end{abstract}

\section{Introduction}

The past year has seen a renewal of interest in end-to-end learning of
communication strategies between pairs of agents represented with deep
networks \cite{Wagner03Emergent}. Approaches of this kind make it possible to learn decentralized
policies from scratch \cite{Foerster16Communication,Sukhbaatar16CommNet}, 
with multiple agents coordinating via learned communication protocol.
More generally, any encoder--decoder
model \cite{Sutskever14NeuralSeq}
can be viewed as implementing an analogous communication protocol, with the input encoding 
playing the role of a message in an artificial ``language'' shared by the encoder and 
decoder \cite{Yu16Reinforcer}.
Earlier work has found that 
under suitable conditions, these protocols acquire simple interpretable lexical
\cite{Dircks99LanguageChange,Lazaridou16Communication}
and sequential structure \cite{Mordatch17Language}, even without 
natural language training data.

One of the distinguishing features of natural language is compositionality: the
existence of operations like negation and coordination that can be applied to
utterances with predictable effects on meaning.  RNN models trained for natural
language processing tasks have been found to learn representations that encode
some of this compositional structure---for example, sentence representations for
machine translation encode explicit features for certain syntactic phenomena
\cite{Shi16MTSyntax} and represent some semantic relationships translationally
\cite{Levy14WordVecs}.
It is thus natural to ask whether these ``language-like'' structures also arise
spontaneously in models trained directly from an environment signal.  Rather
than using language as a form of supervision, we propose to use it as a
\emph{probe}---exploiting post-hoc statistical correspondences between natural
language descriptions and neural encodings to discover regular structure in
representation space.

\begin{figure}
\centering
\strut\\[-2em]
\includegraphics[width=\columnwidth,trim=0 3.5in 4.5in 0,clip]{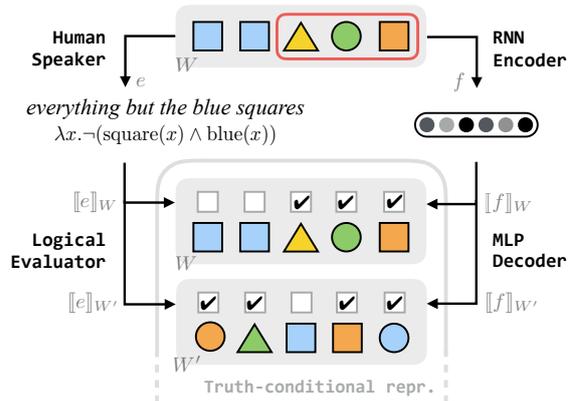}\\[-1em]
\caption{
Overview of our task. Given a dataset of referring expression games, example
human expressions, and their associated logical forms,
we compute explicit denotations both for the original
task and in other possible tasks---giving rise to a truth-conditional representation of the
natural language. We train a recurrent encoder--decoder model to solve the same tasks directly, 
and use the decoder to generate comparable truth-conditional representations of neural encodings.
}
\label{fig:task}
\vspace{-2mm}
\end{figure}

To do this, we need to find (vector, string) pairs with matching semantics,
which requires first aligning unpaired examples of human--human communication with network
hidden states. This is similar to the problem of ``translating'' RNN representations 
recently investigated in \newcite{Andreas17Neuralese}. Here we build on that approach in order to 
perform a detailed analysis of  \emph{compositional} structure in learned ``languages''. 
We investigate a communication game  previously studied by \newcite{FitzGerald13Referring}, 
and make two discoveries:
in a model trained without any access to language data,
\begin{enumerate}
  \item The strategies employed by human speakers in a given communicative
        context are surprisingly good predictors of RNN behavior in the same
        context: humans and RNNs send messages whose interpretations agree on
        nearly 90\% of object-level decisions, even outside the contexts in
        which they were produced.
    \item 
        Interpretable language-like structure naturally arises in the space of
        representations.  We identify geometric regularities corresponding to
        negation, conjunction, and disjunction, and show that it is possible to
        linearly transform representations in ways that approximately correspond
        to these logical operations.
\end{enumerate}
\section{Task}

We focus our evaluation on a communication game due to
\newcite{FitzGerald13Referring} (\autoref{fig:task}, top). In this game, the
\emph{speaker} observes (1) a \emph{world} $\world$ of 1--20 objects labeled
with with attributes and (2) a designated \emph{target} subset $\target$ of
objects in the world. The \emph{listener} observes only $\world$, and the
speaker's goal is to communicate a representation of $\target$ that enables the
listener to accurately reconstruct it. The \genx dataset collected for this
purpose contains 4170 human-generated natural-language referring expressions and
corresponding logical forms for 273 instances of this game. Because these
human-generated expressions have all been pre-annotated, we treat language and
logic interchangeably and refer to both with the symbol $\nl$. We write
$\nl(\world)$ for the expression generated by a human for a particular world
$\world$, and $\denote{\nl}_\world$ for the result of evaluating the logical
form $\nl$ against $\world$.

We are interested in using language data of this kind to analyze the behavior of
a deep model trained to play the same game. We focus our analysis on a standard
RNN encoder--decoder, with the encoder playing the role of the speaker and the
decoder playing the role of the listener. The encoder is a single-layer RNN with
GRU cells \cite{Cho14GRU} that consumes both the input world and target labeling and outputs a
64-dimensional hidden representation. We write $\rl(\world)$ for the output of
this encoder model on a world $W$.  To make predictions, this representation is
passed to a decoder implemented as a multilayer perceptron.  The decoder makes
an independent labeling decision about every object in $\world$ (taking as input
both $\rl$ and a feature representation of a particular object $\world_i$).  We
write $\denote{\rl}_\world$ for the full vector of decoder outputs on $\world$.
We train the model maximize classification accuracy on randomly-generated scenes
and target sets of the same form as in the \genx dataset.

\section{Approach}
\label{sec:approach}

We are not concerned with the RNN model's raw performance on this task (it
achieves nearly perfect accuracy). Instead, our goal is to explore what kinds of
messages the model computes in order to achieve this accuracy---and specifically
whether these messages contain high-level semantics and low-level structure
similar to the referring expressions produced by humans.  But how do we judge
semantic equivalence between natural language and vector representations?  Here,
as in \newcite{Andreas17Neuralese}, we adopt an approach inspired by formal
semantics, and represent the meaning of messages via their \emph{truth
conditions} (\autoref{fig:task}).

For every problem instance $\world$ in the dataset, we have access to one or
more human messages $\nl(\world)$ as well as the RNN encoding $\rl(\world)$. The
truth-conditional account of meaning suggests that we should judge $\nl$ and
$\rl$ to be equivalent if they designate the same set of of objects in the world
\cite{Davidson67Truth}.  But it is not enough to compare their predictions
solely in the context where they were generated---testing if
$\denote{\nl}_\world = \denote{\rl}_\world$---because any pair of models that
achieve perfect accuracy on the referring expression task will make the same
predictions in this initial context, regardless of the meaning conveyed.

Instead, we sample a collection of alternative worlds $\{\world_i\}$ observed
elsewhere in the dataset, and compute a tabular meaning representation
$\rep{\nl} = \{\denote{\nl}_{\world_i}\}$ by evaluating $\nl$ in each world
$\world_i$.  We similarly compute $\rep{\rl} = \{\denote{\rl}_{\world_i}\}$,
allowing the learned decoder model to play the role of logical evaluation for
message vectors.
For logically equivalent messages, these tabular representations are guaranteed
to be identical, so the sampling procedure can be viewed as an approximate test
of equivalence. It additionally allows us to compute softer notions of
equivalence by measuring agreement on individual worlds or objects.

\section{Interpreting the meaning of messages}

We begin with the simplest question we can answer with this tool: how often do
the messages generated by the encoder model have the same meaning as messages
generated by humans for the same context?  Again, our goal is not to evaluate
the performance of the RNN model, but instead our ability to understand its
behavior. Does it send messages with human-like semantics? Is it more explicit?
Or does it behave in a way indistinguishable from a random classifier?

\begin{figure}[b!]
  \centering
  \includegraphics[width=.9\columnwidth,trim=0 2.8in 5in 0,clip]{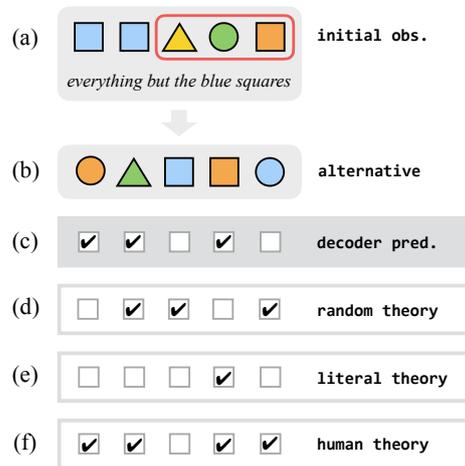}
  \caption{Evaluating theories of model behavior. First, the encoder is run on
    an initial world (a), producing a representation whose meaning we would like
    to understand (see \autoref{fig:task}). We then observe the behavior of the
    decoder holding this representation fixed but replacing the underlying world
    representation with alternatives like (b).
    We compare the true decoder output to a number of
    \emph{theories} of its behavior.  The random theory (d) outputs a random
    decision for every object.  The literal theory (e) predicts that the decoder
    will output a positive label only on those objects that exactly match some
    object in the initial observation. The human theory (f) assigns labels
    according to the logical semantics of the utterance produced by a human
    presented with the initial observation.
    }
    \label{fig:theories}
    \vspace{-1mm}
\end{figure}

For each scene in the \genx test set, we compute the model-generated message $\rl$ and its
tabular representation $\rep{\rl}$, and measure the extent to which this agrees with representations
produced by three ``theories'' of model behavior (\autoref{fig:theories}):
  (1) a \textbf{random} theory that accepts or rejects objects with uniform probability,
  (2) a \textbf{literal} theory that predicts membership only for objects that exactly match
  some object in the original target set, and
  (3) a \textbf{human} theory that predicts according to the most frequent logical form associated
  with natural language descriptions of the target set (as described in the
  preceding section).
We evaluate agreement at the level of individual objects, worlds, and full tabular meaning
representations.

Results are shown in \autoref{tab:corresp}. Model behavior is well explained by
human decisions in the same context: object-level decisions can be predicted
with close to 90\% accuracy based on human judgments alone, and a third of
message pairs agree exactly in every sampled scene, providing strong evidence
that they carry the same semantics.

\begin{table}
\footnotesize
\centering
\begin{tabular}{ccccc}
\toprule
& Theory & Objects & Worlds & Tables \\
\midrule
\multirow{3}{*}{\rot{\footnotesize All}} 
& Random  & 0.50    & 0.00   & 0.00 \\
& Literal & 0.74    & 0.27   & 0.05 \\
& Human   & 0.92    & 0.63   & 0.35 \\
\bottomrule
\end{tabular}
\caption{
  Agreement with predicted model behavior for the high-level semantic
  correspondence task, computed for objects (single entries in tabular
  representation), worlds (rows), and full tables.  Referring expressions $\nl$
  generated by humans in a single communicative context are highly predictive of
  how learned representations $\rl$ will be interpreted by the decoder across
  multiple contexts. 
}
\label{tab:corresp}
\end{table}

These results suggest that the model has learned a communication 
strategy that is at least superficially language-like: it admits
representations of the same kinds of communicative abstractions that
humans use, and makes use of these abstractions with some frequency.
But this is purely a statement about the high-level
behavior of the model, and not about the structure of the space
of representations. Our primary goal
is to determine whether this behavior is achieved using low-level
structural regularities in vector space that can themselves be
associated with aspects of natural language communication. 

\section{Interpreting the structure of messages}
\label{sec:comp}

For this we turn to a focused investigation of three specific logical constructions
used in natural language: a unary operation (negation) and two binary operations
(conjunction and disjunction). All are used in the training data, with a variety of
scopes (e.g.\ \emph{all green objects that are not a triangle},
\emph{all the pieces that are not tan arches}). 

Because humans often find it useful
to specify the target set by exclusion rather than inclusion, we first hypothesize
that the RNN language might find it useful to incorporate some mechanism 
corresponding to negation, and that messages can be predictably ``negated'' in vector space.
To test this hypothesis, we first collect examples of the form $(\nl, \rl, \nl', \rl')$,
where $\nl' = \lnot \nl$, $\rep{\nl} = \rep{\rl}$, and $\rep{\nl'} = \rep{\rl'}$.
In other words, we find pairs of pairs of RNN representations $\rl$ and $\rl'$ for
which the natural language messages $(\nl, \nl')$ serve as a denotational
\emph{certificate} that $\rl'$ behaves as a negation of $\rl$.
If the learned
model does not have any kind of primitive notion of negation,
we expect that it will not be possible to find any kind of predictable relationship
between pairs $(\rl, \rl')$. (As an extreme example, we could imagine every possible
prediction rule being associated with a different point in the representation space,
with the correspondence between position and behavior essentially random.) Conversely, if there is a first-class notion of negation,
we should be able to select an arbitrary representation vector $\rl$ with an associated referring
expression $\nl$, apply some transformation $N$ to $\rl$, and be able to predict 
\emph{a priori} how the decoder model will interpret the representation $N \rl$---i.e.\ 
in correspondence with $\lnot \nl$. 

\begin{table}
\centering
\footnotesize
\begin{tabular}{ccccc}
\toprule
& Theory & Objects & Worlds & Tables \\
\midrule
\multirow{3}{*}{\rot{\footnotesize Neg.}} 
& Random   & 0.50    & 0.00   & 0.00 \\
& Literal  & 0.50    & 0.12   & 0.03 \\
& Negation & 0.97    & 0.81   & 0.45 \\
\midrule
\multirow{3}{*}{\rot{\footnotesize Disj.}} 
& Random      & 0.50   & 0.00   & 0.00 \\
& Literal     & 0.58   & 0.09   & 0.01 \\
& Disjunction & 0.92   & 0.54   & 0.19 \\
\midrule
\multirow{3}{*}{\rot{\footnotesize Conj.}} 
& Random      & 0.50    & 0.00   & 0.00 \\
& Literal     & 0.81    & 0.19   & 0.01 \\
& Conjunction & 0.90    & 0.56   & 0.37 \\
\bottomrule
\end{tabular}
\caption{
  Agreement with predicted model behavior for negation, conjunction, and disjunction tasks (top 
  to bottom). 
  Evaluation is performed on transformed message vectors as described in \autoref{sec:comp}.
  We discover a robust linear transformation of message vectors corresponding 
  to negation, as well as evidence of structured representations of binary 
  operations.
}
\label{tab:logic}
\end{table}

\begin{figure}
  \vspace{-4mm}
\centering
\raisebox{5.5em}{(a)}~~~\includegraphics[width=0.85\columnwidth,trim=0 1in 0 0,clip]{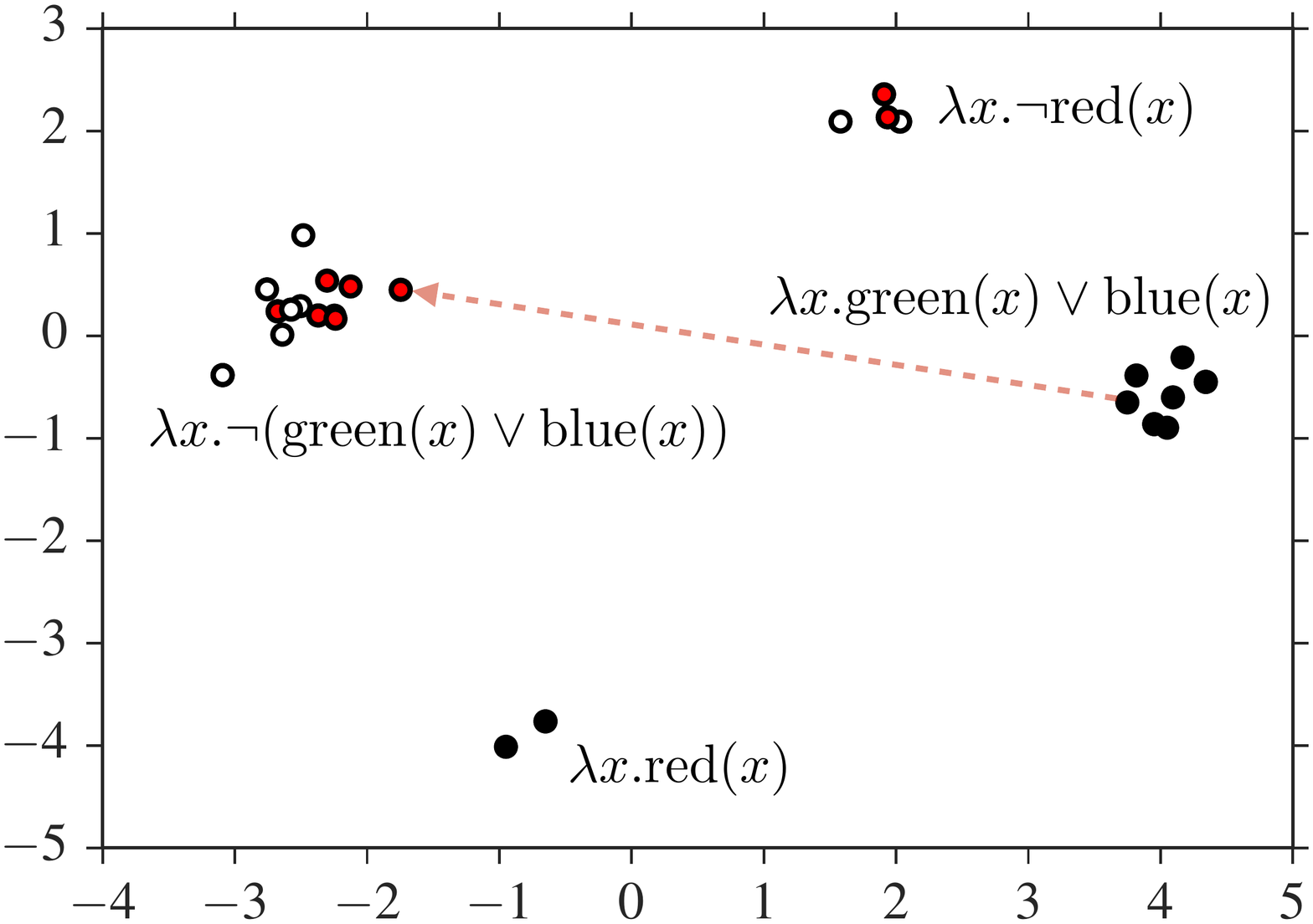} \\
\raisebox{5.5em}{(b)}~~~\includegraphics[width=0.85\columnwidth,trim=0 1in 0 0,clip]{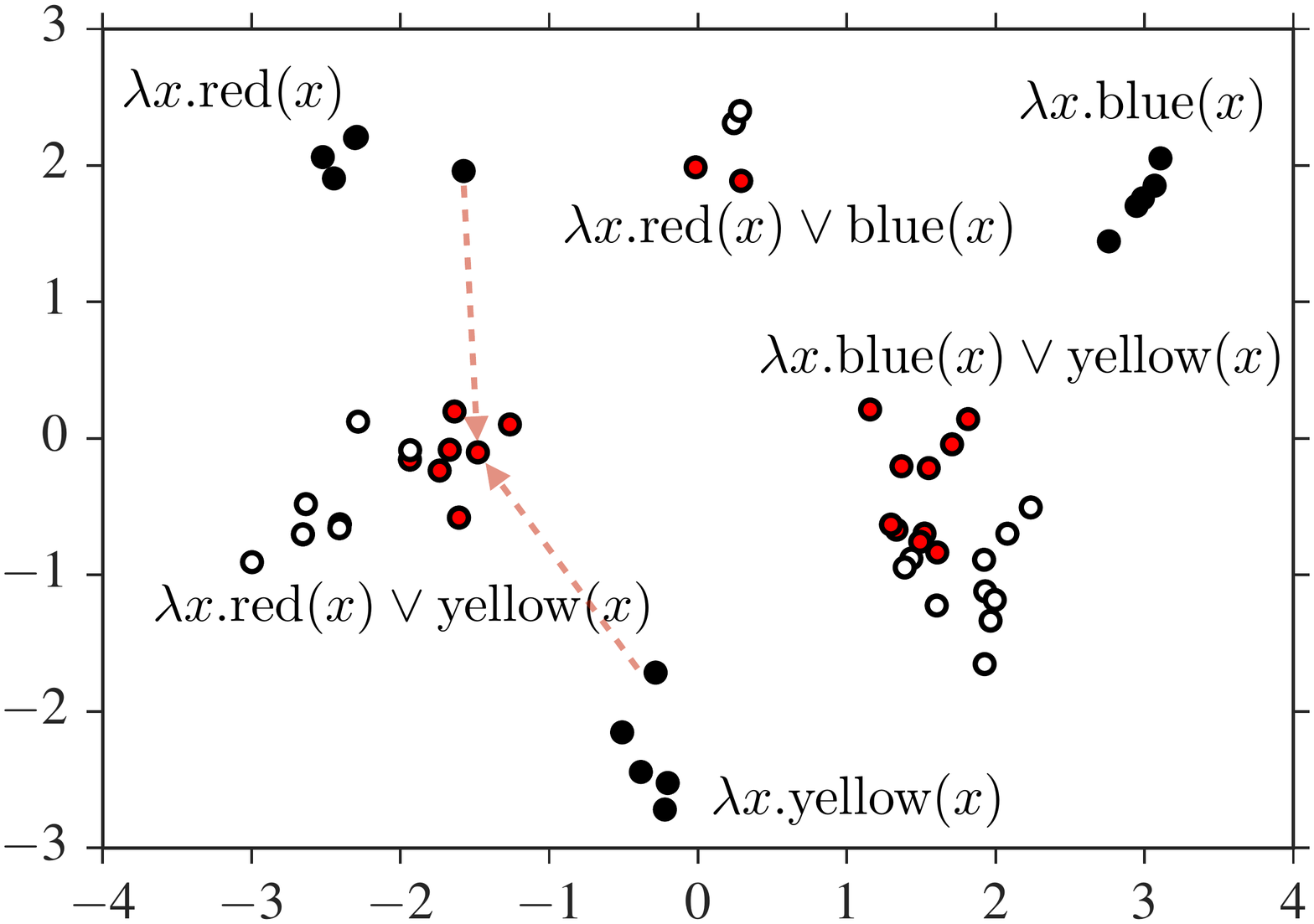} \\
\caption{
Principal components of structured message transformations discovered by our experiments.
(a) Negation: black and white dots show raw message vectors denotationally equivalent to the provided
logical cluster label (\autoref{sec:approach}). Red dots show the result of transforming black dots with the estimated 
negation operation $N$.
(b) The corresponding experiment for disjunction using the transformation $M$.
}
\label{fig:vis}
\end{figure}

Here we make the strong assumption that the negation operation is not only predictable but
\emph{linear}. 
Previous work has found that linear operators are powerful enough to capture many
hierarchical and relational structures \cite{Paccanaro02LRE,Bordes14GraphEmbedding}.
Using examples $(\rl, \rl')$ collected from the training set as described above,
we compute the least-squares estimate
$ \hat{N} = \argmin_N \sum ||N \rl - \rl'||_2^2\ . $
To evaluate, we collect example representations from the test set that are equivalent to known
logical forms, and measure how frequently model behaviors $\rep{N \rl}$ agree with the logical
predictions $\rep{\lnot \nl}$---in other words, how often the linear operator $N$ actually corresponds
to logical negation. Results are shown in the top portion of \autoref{tab:logic}. Correspondence with the
logical form is quite high, resulting in 97\% agreement at the level of individual objects 
and 45\% agreement on full representations. We conclude that the estimated linear operator 
$\hat{N}$ is analogous to negation in natural language. Indeed, the behavior of this operator
is readily visible in \autoref{fig:vis}: 
predicted negated forms (in red) lie close in vector space to their true values, and
negation corresponds roughly to mirroring across a central point.

In our final experiment, we explore whether the same kinds of linear maps can
be learned for the binary operations of conjunction and disjunction. As in the previous
section, we collect examples from the training data of representations whose denotations 
are known to correspond to groups of logical forms in the desired relationship---in 
this case tuples
$(\nl, \rl, \nl', \rl', \nl'', \rl'')$, where $\rep{\nl} = \rep{\rl}$, $\rep{\nl'} = 
\rep{\rl'}$, $\rep{\nl''} = \rep{\rl''}$ and either $\nl'' = \nl \land \nl'$ (conjunction) 
or  $\nl'' = \nl \lor \nl'$ (disjunction).
Since we expect that our operator will be 
symmetric in its arguments, we solve for
$ \hat{M} = \argmin_M \sum ||Mf + Mf' - f''||_2^2 .$

Results are shown in the bottom portions of \autoref{tab:logic}. Correspondence
between the behavior predicted by the contextual logical form and the model's
actual behavior is less tight than for negation.  At the same time, the
estimated operators are clearly capturing some structure: in the case of
disjunction, for example, model interpretations are correctly modeled by the
logical form 92\% of the time at the object level and 19\% of the time at the
denotation level. This suggests that the operations of conjunction and
disjunction do have some functional counterparts in the RNN language, but that
these functions are not everywhere well approximated as linear.

\section{Conclusions}

Building on earlier tools for identifying neural codes with natural language
strings, we have presented a technique for exploring compositional structure in
a space of vector-valued representations. Our analysis of an encoder--decoder
model trained on a reference game identified a number of language-like
properties in the model's representation space, including transformations
corresponding to negation, disjunction, and conjunction. One major question left
open by this analysis is what happens when multiple transformations are applied
hierarchically, and future work might focus on extending the techniques in this
paper to explore recursive structure. We believe our experiments so far
highlight the usefulness of a denotational perspective from formal semantics
when interpreting the behavior of deep models.

\bibliography{jacob}
\bibliographystyle{emnlp_natbib}

\end{document}